\documentclass{segabs}

\usepackage{amsmath,graphicx}

\usepackage{booktabs} 
\usepackage{tabularx}
\usepackage{enumitem}
\usepackage{cite}
\usepackage{ifpdf}
\usepackage{algorithmic}
\usepackage{multirow}
\begin{document}

\twocolumn[{%
\vspace{30mm}
\large
\begin{itemize}[leftmargin=2.5cm, align=parleft, labelsep=2cm, itemsep=4ex]

\item[\textbf{Citation}]{K. Kokilepersaud and G. AlRegib, "Volumetric Supervised Contrastive Learning for Seismic Semantic Segmentation," 2020 27th \textit{The International Meeting for Applied Geoscience \& Energy}, Houston, Texas, 2022.}

\item[\textbf{Review}]{Date of Publication: May 17, 2022}

\item[\textbf{Bib}]  {@inproceedings\{kokilepersaud2022volumetric,\\
    title=\{Volumetric Supervised Contrastive Learning for Seismic Semantic Segmentation\},\\
    author=\{Kokilepersaud, Kiran, and Prabhushankar, Mohit and AlRegib, Ghassan\},\\
    booktitle=\{The International Meeting for Applied Geoscience \& Energy\},\\
    year=\{2022\}\}}
\item[\textbf{Contact}]{
\{kpk6, mohit.p, alregib\}@gatech.edu}

\item[\textbf{URL}]{
https://ghassanalregib.info/}

\end{itemize}
}]

\newpage
\clearpage
\setcounter{page}{1}
\title{Volumetric Supervised Contrastive Learning for Seismic Semantic Segmentation}

\renewcommand{\thefootnote}{\fnsymbol{footnote}} 

\author{Kiran Kokilepersaud, Mohit Prabhushankar, Ghassan AlRegib, Center for Energy and Geo Processing (CeGP), Georgia Institute of Technology}

\maketitle

\begin{abstract}
In seismic interpretation, pixel-level labels of various rock structures can be time-consuming and expensive to obtain. As a result, there oftentimes exists a non-trivial quantity of unlabeled data that is left unused simply because traditional deep learning methods rely on access to fully labeled volumes. To rectify this problem, contrastive learning approaches have been proposed that use a self-supervised methodology in order to learn useful representations from unlabeled data.  However, traditional contrastive learning approaches are based on assumptions from the domain of natural images that do not make use of seismic context. In order to incorporate this context within contrastive learning, we propose a novel positive pair selection strategy based on the position of slices within a seismic volume. We show that the learnt representations from our method out-perform a state of the art contrastive learning methodology in a semantic segmentation task.
\end{abstract}

\section{Introduction}
During exploration for oil and gas, seismic acquisition technology outputs a large amount of data in order to obtain 2D and 3D images of the surrounding subsurface layers. Despite the potential advantages that come with access to this huge quantity of data, processing and subsequent interpretation remains a major challenge for these companies \citep{mohammadpoor2020big}. Interpretation of seismic volumes is done in order for geophysicists to identify relevant rock structures in regions of interest. Conventionally, these structures are identified and labeled by trained interpreters, but this process can be expensive and labor intensive. This results in the existence a large amount of unlabeled data alongside a smaller number that has been fully interpreted. 

\begin{figure}[t]
\centering
\includegraphics[width=\columnwidth]{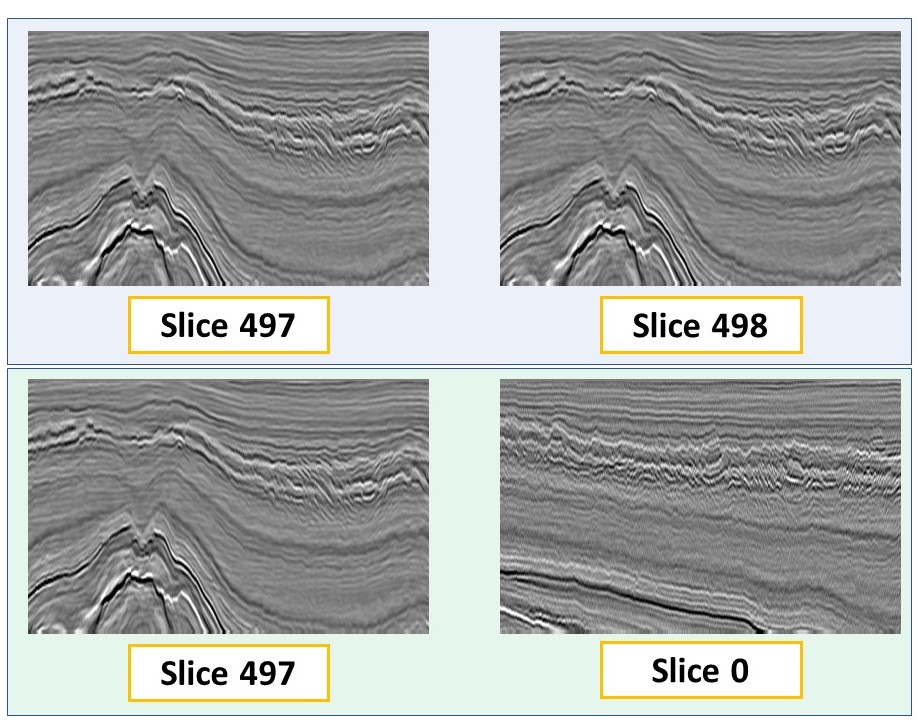}

\caption{This figure shows seismic images from different regions of the Netherlands F3 Block. It is clear that images that come from adjacent slices in the volume have more structural features in common than those that are far apart in the volume. \vspace{-.3cm}}

\label{fig: similar}
\end{figure}

To overcome these issues, work has gone into using deep learning \citep{di2018deep} to automate the interpretation process. However, a major problem with any conventional deep learning setup  is the dependence on having access to a large pool of training data. As discussed, this dependency is not reliable within the context of seismic. In order to overcome this reliance on labeled data as well as leverage the potentially larger amount of unlabeled data, contrastive learning \citep{le2020contrastive} has emerged as a promising research direction. The goal of contrastive learning approaches is to learn distinguishing features of data without needing access to labels. This is done through algorithms that learn to to associate images with similar features (positives) together and disassociate images with differing features (negatives) apart. Traditional approaches, such as \citep{chen2020simple}, do this by taking augmentations from a single image and treating these augmentations as the positives, while all other images in the batch are treated as the negative pairs. These identified positive and negative pairs are input into a contrastive loss that minimizes the distance between positive pairs of images and maximizes the distance between negative pairs in a lower dimensional space. These approaches work well within the natural image domain, but may exhibit certain flaws within the context of seismic imaging. For example, naive augmentations could potentially distort the textural elements that constitute different classes of rock structures. A better approach for identifying positive pairs of images would be by considering the position of instances within the volume. We can observe this in Figure \ref{fig: similar} where seismic images that exist closer to each other in a volume exhibit more structural components in common than those that are further apart. Therefore, these images that are closer to each other within a volume have similar features that a contrastive loss would be able to distinguish from features of other classes of rocks. 

In this work, we propose to take advantage of the correlations between images close to each other in a volume through a contrastive learning methodology. Specifically, we partition a seismic volume during training into smaller subsets and assign the slices of each subset the same volume based label.  We utilize these volume based labels to train an encoder network with a supervised contrastive loss \citep{khosla2020supervised}. Effectively this means that the model is trained to learn to associate images close in the volume together and disassociate images that are further apart. From the representation space learnt by training in this manner, we fine-tune an attached semantic segmentation head using the available ground truth labels. The proposed contributions of this paper are:

\begin{enumerate}
  \item We introduce contrastive learning within the context of a seismic semantic segmentation task. 
  \item We introduce a novel positive pair selection strategy for seismic on the basis of generated volumetric position labels.
\end{enumerate}

\section{Related Works}
The original usage of deep learning for seismic interpretation tasks was within the context of supervised tasks \citep{di2018deep} where the authors performed salt-body delineation. Further work into supervised tasks included semantic segmentation using deconvolution networks \citep{alaudah2019machine}. Deep learning was also utilized for the task of acoustic impedance estimation \citep{mustafa2020spatiotemporal,mustafa2020joint}.  However, it was quickly recognized that labeled data is expensive and training on small datasets leads to poor generalization of seismic models. For this reason, the research focus switched to methods without as high of a dependence on access to a large quantity of labeled data. This includes \citep{alaudah2017weakly, alaudah2019facies, alaudah2017weakly_non} where the authors introduced various methods based on weak supervision of structures within seismic images. Other work introduced semi-supervised methodologies such as \citep{alfarraj2019semisupervised} for the task of elastic impedance inversion. \citep{lee2018automatic} introduced a labeling strategy that made use of well logs alongside seismic data.  \citep{shafiq2018leveraging} and \citep{shafiq2018towards} introduced the idea of leveraging learnt features from the natural image domain. Related work \citep{shafiq2022novel} and \citep{shafiq2018role} showed how saliency could be utilized within seismic interpretation. More recent work involves using strategies such as explainability \citep{prabhushankar2020contrastive} and learning dynamics analysis \citep{benkert2021explainable}. 

Despite the potential of pure self-supervised approaches, there isn't a significant body of work within the domain of seismic. Work such as \citep{aribido2020self} and \citep{aribido2021self} showed how structures can be learnt in a self-supervised manner through manipulation of a latent space. \citep{soliman2020s} created a self and semi-supervised methodology for seismic semantic segmentation.  More recent work \citep{huang2022self} introduced a strategy to reconstruct missing data traces.  The most similar work to ours occurs within the medical field where \citep{zeng2021positional} uses a contrastive learning strategy based on slice positions within an MRI and CT setting. Our work differs from previous ones due to the introduction of a contrastive learning strategy based on volume positions within a seismic setting.

\begin{figure*}[t]
\centering
\includegraphics[scale = .5]{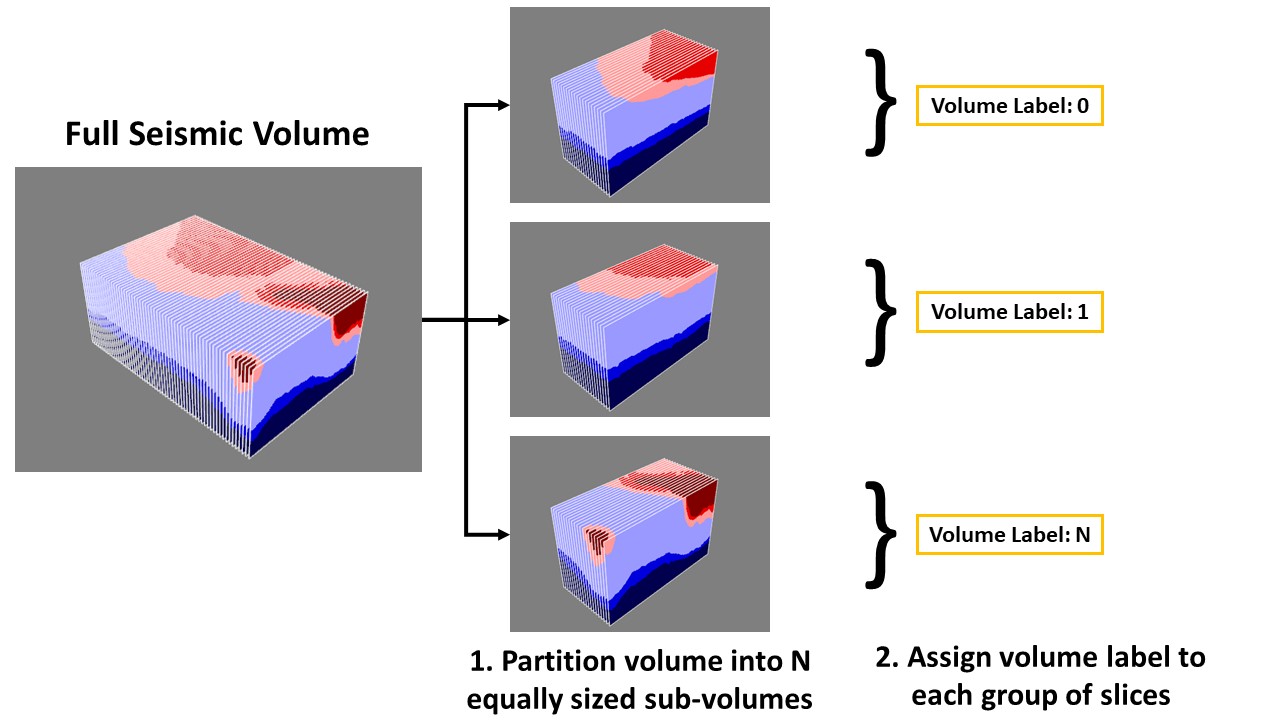}

\caption{Volume label generation process. 1) Divide the original volume into N equally sized sub-volumes. 2) Assign each slice within each sub-volume the same volume label. \vspace{-.3cm}}

\label{fig: volume}
\end{figure*}
\section{Methodology}
\subsection{Dataset}

For all of our experiments, we utilize the publicly available F3
block located in the Netherlands \citep{alaudah2019machine}. This dataset contains full semantic segmentation annotations of the rock structures present. We utilize the training and test sets introduced by the original author. The training volume consists of 400 in-lines and 700 cross-lines. We utilize the 700 cross-lines during training. The test set includes data from two neighboring volumes. This first volume consists of 600 labeled in-lines and 200 labeled cross-lines. The second volume has 200 in-lines and 700 cross-lines. For testing, we combine the cross-lines from each volume to form a larger 900 cross-line test set. These 900 images are divided into three test splits consisting of 300 images each. The results shown are the average mean intersection over union across each of these test splits.

\subsection{Volume Based Labels}

In order to select better positive pairs for a contrastive loss, we propose to assign pseudo-labels to cross-lines based on their position within the volume. This process is shown in Figure \ref{fig: volume}. From our starting set of 700 cross-lines, we define a hyper-parameter $N$ that dictates the number of equally sized partitions that we will divide the volume into. For example, if $N$ = 100 then the volume will be divided into 100 equally sized partitions consisting of 7 cross-lines each. After dividing the volume in this manner, cross-lines belonging to the same sub volume are assigned the same volume position label $V_{L}$. This volume label acts to identify cross-lines that are more likely to share structural features in common due to being next to each other within the F3 block.
\begin{figure*}[t]

\centering
\includegraphics[scale = .5]{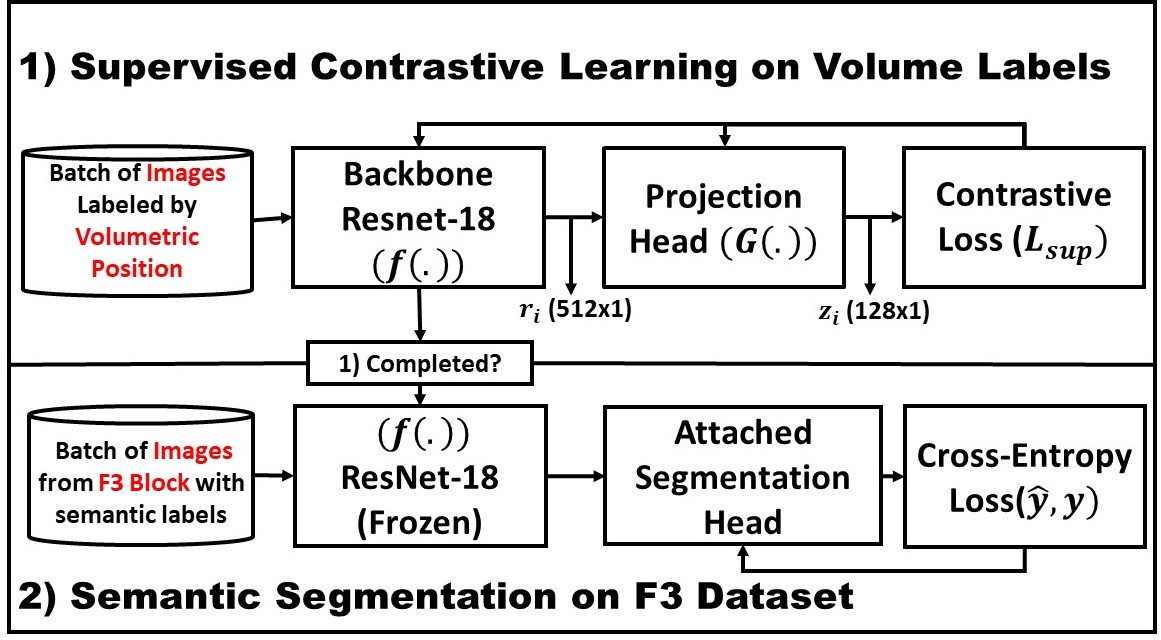}

\caption{Seismic volumetric contrastive learning overall approach. 1) Supervised contrastive learning by assigning labels based on position within F3 Block. 2) Use representations learnt from this pre-training to fine-tune for the semantic segmentation task. \vspace{-.3cm}}

\label{fig: overview}
\end{figure*}

\subsection{Supervised Contrastive Learning Framework}

Once we have the volume labels ($V_{L}$), we utilize the supervised contrastive loss \citep{khosla2020supervised} to bring embeddings of images with the same volume label together and push apart embeddings of images with differing volume labels. The overall setup is detailed through the flowchart shown in Figure \ref{fig: overview}. We first pre-train a backbone ResNet-18 model \citep{he2016deep} using the volume labels to identify positive and negative pairs of images for the supervised contrastive loss. Each cross-line image $x_{i}$ is passed through the ResNet-18 encoder network $f(\cdot)$, producing a $512\times 1$ dimensional vector $r_{i}$. This vector is further compressed through a projection head $G(.)$ which is set to be a multi-layer perceptron with a single hidden layer. This projection head is used to reduce the dimensionality of the representation and is discarded after training. The output of $G(.)$ is a $128\times 1$ dimensional embedding $z_{i}$. In this embedding space, the dot product of images with the same volume label (the positive samples) are maximized and those with different volume labels (the negative samples) are minimized. This takes the form of equation \ref{eq:sup} where positive instances for image $x_{i}$ come from the set $P(i)$ and positive and negative instances come from the set $A(i)$. $z_{i}$ and $z_{p}$ are embeddings that originated from each of these sets respectively. $\tau$ is a temperature scaling parameter set to .07 for all experiments. 

\begin{equation}
    \label{eq:sup}
     L_{sup} = \sum_{i\in{I}} \frac{-1}{|P(i)|}\sum_{p\in{P(i)}}log\frac{exp(z_{i}\cdot z_{p}/\tau)}{\sum_{a\in{A(i)}}exp(z_{i}\cdot z_{a}/\tau)}
\end{equation}

After pre-training the network via the supervised contrastive loss on volume position labels, we move to step two in the methodology detailed in Figure \ref{fig: overview}. In this step, the weights of the previously trained encoder are frozen and a semantic segmentation head from the Deep Lab v3 architecture \citep{chen2018encoder} is appended to the output of the encoder. We pass batches of images from the same 700 cross-lines that we used in the previous step, but we now re-introduce the associated semantic segmentation labels for each cross-line. The output of the head is a pixel-level probability vector map $\hat{y}$ that is used as input to a cross-entropy loss with the ground truth segmentation labels $y$. This loss is used to train the segmentation head to segment the volume into relevant rock structure regions. In this way, we fine-tune the semantic segmentation head using the representations learnt from the contrastive loss.

\section{Results}
Our goal is to see how the representations learnt from our contrastive learning strategy perform relative to representations learnt from other methods. In order to do this we compare against a state of the art contrastive learning algorithm SimCLR \citep{chen2020simple}. The architecture was kept constant as ResNet-18 for both experiments. Augmentations for both methods during the contrastive training step involved random resize crops to a size of 224, horizontal flips, color jittering, and normalization to the mean and standard deviation of the seismic dataset. During training of the segmentation head, augmentations were limited to just normalization of the data. The batch size was set to 64. Training was performed for 50 epochs in for both the contrastive pre-training as well as the segmentation head fine-tuning. A stochastive gradient descent optimizer was utilized with a learning rate of .001 and a momentum of .9. We assess the quality of our method through the average mean intersection over union metric of the three test splits we introduced. 

A summary of our results is displayed in Table \ref{tab: miou}. It can be observed that regardless of the number of partitions that we divide the volume into, our method out-performs the state of the art SimCLR framework. It is also interesting to analyze the effect of varying the partition hyper-parameter $N$. With a lower value of N, this means that the volume is split into fewer number of distinct partitions. As a result, this means that there exists more images within each partition. However, the exact opposite scenario occurs when $N$ is a larger value. In this case, there are more partitions and each partition contains a fewer number of images. The advantage of this is that the images within each partition are more likely to contain similar structural features in common that the model can learn to distinguish. This is because, with a higher number of partitions, the contrastive loss is effectively choosing positive pairs from images that exist very close to each other within the volume and; hence, exhibit stronger correlations with each other. This acts as a preliminary explanation to the observed performance increase shown in Figure \ref{tab: miou} when going from $N$ = 10 to $N$ = 150. 

\begin{table}[]
\centering
\begin{tabular}{@{}cc@{}}

\toprule
\multicolumn{2}{c}{Semantic Segmentation Results}                       \\ \midrule
\multicolumn{1}{|c|}{Method} & \multicolumn{1}{c|}{MIOU} \\ \midrule

SimCLR & .6781 \\
\bottomrule
Volume Labels (N=10)  &  .6862 \\
Volume Labels (N=100) & .6860 \\ 
Volume Labels (N=150) &  \textbf{.6874} \\

 \bottomrule
\end{tabular}
\caption{We report average MIOU performance across three different test sets for our method and the SimCLR framework. N refers to the number of partitions that the training set volume was divided into before applying the proposed method.}
\label{tab: miou}
\end{table}
\section{Conclusion}
Labels are time consuming and expensive to obtain within the seismic domain, due to a reliance on an expert interpreter. For this reason, contrastive learning presents an opportunity because it provides a means by which unlabeled data can be incorporated into the process of training a model for semantic segmentation of rock volumes. It works by defining positive and negative pairs of images to utilize within a contrastive loss. We show in this paper that one novel approach to choose positives is to assign positional labels to cross-lines that are adjacent to each other within a seismic volume. From these assigned labels, we can then use a supervised contrastive loss in order to train an encoder network to learn distinguishing characteristics of seismic data from a contrastive loss. Training in this manner led to a representation space more consistent with the seismic setting and was shown to out-perform a state of the art self-supervised methodology in a semantic segmentation task.
\twocolumn

\bibliographystyle{seg}  
\bibliography{ref}

\end{document}